\documentclass[letter, 10pt, conference]{ieeeconf}      
\IEEEoverridecommandlockouts  

\usepackage{graphicx}
\usepackage{amsmath}
\usepackage{tikz}
\usepackage{import}
\usepackage{xspace}
\usepackage{caption}
\usepackage{subcaption}
\usepackage{booktabs,multirow,multicol}
\let\labelindent\relax
\usepackage{enumitem}
\usepackage{url}
\setitemize{noitemsep,topsep=0pt,parsep=0pt,partopsep=0pt}

\renewcommand{\baselinestretch}{0.985}
\definecolor{CommentRed}{rgb}{0.7,0,0}
\definecolor{CommentGreen}{rgb}{0,0.7,0}
\definecolor{CommentBlue}{rgb}{0,0,0.7}
\definecolor{CommentJeff}{rgb}{0,0.7,0.7}

\newcommand*{\eg}{e.g.\@\xspace}

\usepackage{amsmath,amsfonts,bm}

\def\eqref#1{equation~\ref{#1}}

\def\1{\bm{1}}

\def\rva{{\mathbf{a}}}

\def\rvc{{\mathbf{c}}}

\def\rvm{{\mathbf{m}}}

\def\rvo{{\mathbf{o}}}

\def\rvx{{\mathbf{x}}}
\def\rvy{{\mathbf{y}}}

\DeclareMathAlphabet{\mathsfit}{\encodingdefault}{\sfdefault}{m}{sl}
\SetMathAlphabet{\mathsfit}{bold}{\encodingdefault}{\sfdefault}{bx}{n}

\def\gA{{\mathcal{A}}}

\def\gD{{\mathcal{D}}}

\def\gL{{\mathcal{L}}}

\def\gO{{\mathcal{O}}}

\title{\LARGE \bf Urban Driving with Conditional Imitation Learning}

\author{
  Jeffrey Hawke, Richard Shen, Corina Gurau, Siddharth Sharma, Daniele Reda, Nikolay Nikolov \\ Przemys\l{}aw Mazur, Sean Micklethwaite, Nicolas Griffiths, Amar Shah, Alex Kendall%
\thanks{The authors are with Wayve, in London, UK. Equal contribution. {\tt\small research@wayve.ai}}%
}

\begin{document}

\maketitle



\begin{abstract}
Hand-crafting generalised decision-making rules for real-world urban autonomous driving is hard.
Alternatively, learning behaviour from easy-to-collect human driving demonstrations is appealing.
Prior work has studied imitation learning (IL) for autonomous driving with a number of limitations.
Examples include only performing lane-following rather than following a user-defined route, only using a single camera view or heavily cropped frames lacking state observability, only lateral (steering) control, but not longitudinal (speed) control and a lack of interaction with traffic.
Importantly, the majority of such systems have been primarily evaluated in simulation - a simple domain, which lacks real-world complexities.
Motivated by these challenges, we focus on learning representations of semantics, geometry and motion with computer vision for IL from human driving demonstrations.
As our main contribution, we present an end-to-end conditional imitation learning approach, combining both lateral and longitudinal control on a real vehicle for following urban routes with simple traffic.
We address inherent dataset bias by data balancing, training our final policy on approximately 30 hours of demonstrations gathered over six months.
We evaluate our method on an autonomous vehicle by driving 35km of novel routes in European urban streets.
\end{abstract}

\section{Introduction}
Driving in complex environments is hard, even for humans, with complex spatial reasoning capabilities.
Urban roads are frequently highly unstructured: unclear or missing lane markings, double-parked cars, narrow spaces, unusual obstacles and other agents who follow the road rules to widely varying degrees.
Driving autonomously in these environments is an even more difficult robotics challenge.
The state space of the problem is large; coming up with a driving policy that can drive reasonably well and safely in a sufficiently wide variety of situations remains an open challenge. 
Additionally, while the action space is small, a good driving policy likely requires a combination of high-level hierarchical reasoning and low-level control.

There are two main established paradigms for solving the problem of autonomous driving: a traditional engineering-based approach and a data-driven, machine-learning approach.
The former performs well in structured driving environments, such as highways or modern suburban developments, where explicitly addressing different scenarios is tractable and sufficient for human-like driving~\cite{Thrun:2006:SRW:1210475.1210482, dickmanns2002development, leonard2008perception}. This approach is more mature, and the focus of commercial efforts today. However, is still unclear if this can scale to deploying fully autonomous vehicles world-wide, in complex urban scenarios with wider variation and visual diversity.

In comparison, data driven methods avoid hand-crafted rules and learns from human driving demonstrations by training a policy that maps sensor inputs, such as images, to control commands or a simple motion plan (e.g., steering and speed).
Importantly, such an approach has greater potential to generalise to the large state space and address variations such as weather, time of day, lighting, surroundings, type of roads and behaviour of external agents. The attraction here is that we have seen learned approaches to many vision tasks outperform hand-crafted approaches. We expect that it will also be possible to do so for learned visual control.


Prior works primarily explore the data-driven approach in the context of imitation learning, but to date address only a subset of the state space~\cite{bojarski_end_2016, pomerleau1989alvinn} or the action space~\cite{amini2018variational}.
In more recent work, \cite{codevilla2018end} learn both steering and speed, conditioned on a route command (e.g. turn left, go straight, etc.), but evaluate the method on a small-sized toy vehicle in structured scenarios with no traffic.
Similarly, \cite{amini2018variational} address control conditioned on a route with GPS localisation, though learn only lateral control (steering) for driving empty roads.


\begin{figure*}[!t]
    \centering
    \includegraphics[width=0.8\textwidth]{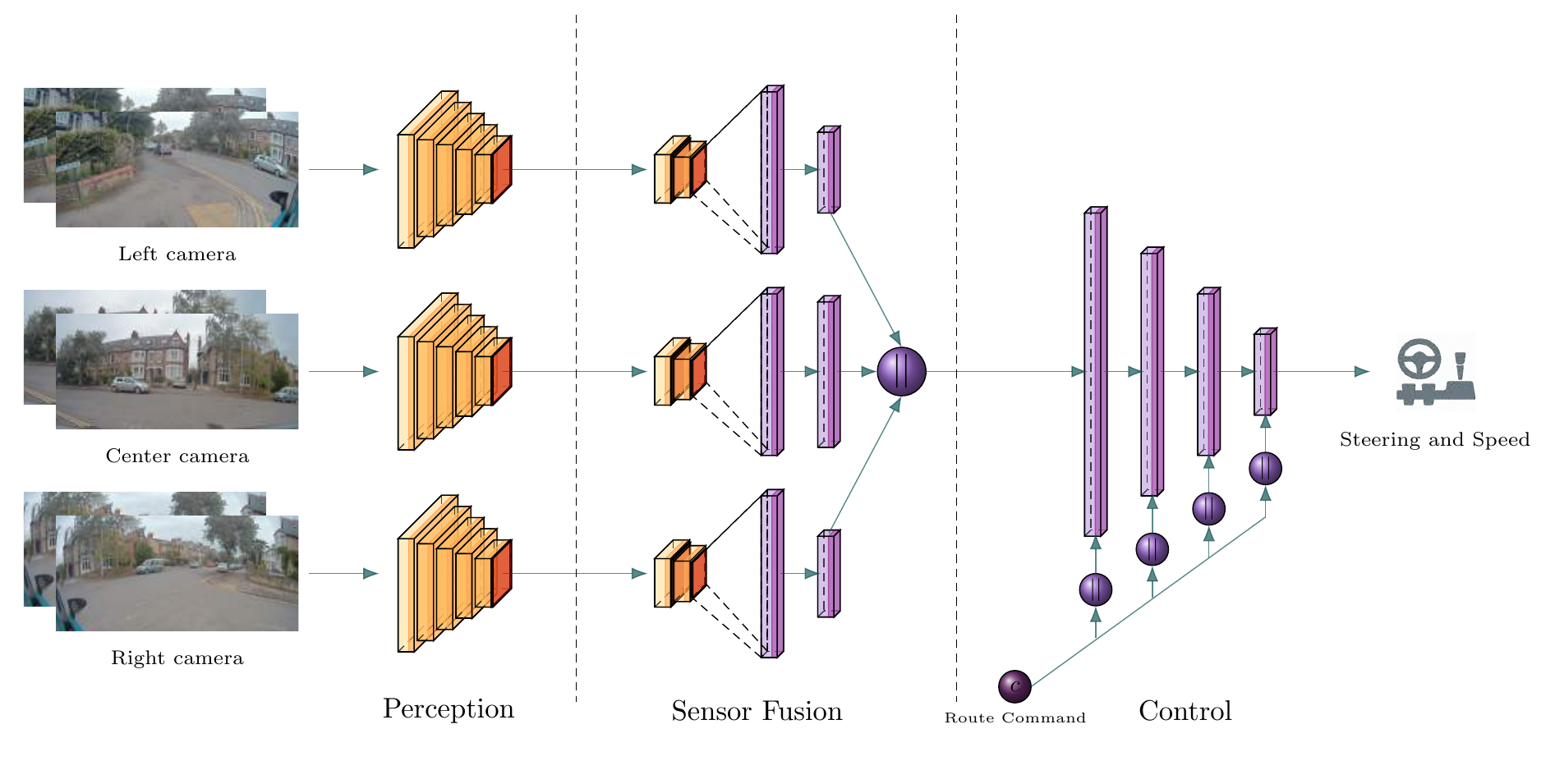}
    \caption{We demonstrate conditional route following using an end-to-end learned driving policy in dense urban driving scenarios, including simple traffic following behaviour. We frame this single fully differentiable architecture conceptually in terms of \emph{perception}, \emph{sensor fusion} and \emph{control} components. The division is purely semantic. Each camera provides an image (or images where using optical flow) to the perception encoders, generating a learned computer vision representation for driving. We generate various perception outputs from this intermediate representation, of semantics, geometry, and motion, shown in Figure \ref{fig:perception_example}. The output features are compressed by sensor fusion and concatenated ($||$ denotes concatenation). Based on this representation and a route command $\rvc$, the network outputs a short motion plan in steering and speed.}
    \label{fig:mcn-architecture}
\end{figure*}

Our main contribution is the development of an end-to-end conditional imitation learning approach, which, to the best of our knowledge, is the first fully learned demonstration of complete control of a real vehicle, following a user-prescribed route in complex urban scenarios. Additionally, we include the first examples of a learned driving policy reacting to other traffic participants. Our method requires monocular camera images as input and can be conditioned on a route command (e.g., taking a turn at an intersection)~\cite{codevilla2018end}, in order to follow the specified route.

Importantly, we utilise small amounts of driving data such that our final policy is trained on only 30 hours of demonstrations collected over six months, yet it generalised to routes it has not been trained to perform.
We train and evaluate\footnote{A video of the learned system driving a vehicle on urban streets is available at \url{https://wayve.ai/blog/learned-urban-driving}} on European urban streets: a challenging, unstructured domain in contrast to simulated environments, or structured motorways and suburban streets.

The novel contributions of this work are:
\begin{itemize}
    \item the first end-to-end learned control policy able to drive an autonomous vehicle in dense urban environments,
    \item a solution to the causal confusion problem, allowing motion information to be provided to the model for both lateral and longitudinal control of the vehicle,
    \item demonstration of data efficiency, showing that it is possible to learn a model capable of decision making in urban traffic with only 30 hours of training data,
    \item comprehensive performance evaluation of end-to-end deep learning policies driving  35km on public roads.
\end{itemize}

\section{Related Work}
%
%
%




The first application of imitation learning (IL) to autonomous driving was ALVINN~\cite{pomerleau1989alvinn}, predicting steering from images and laser range data.
Recently, IL for autonomous driving has received renewed interest due to advances in deep learning.
\cite{bojarski_end_2016, bojarski2017explaining} demonstrated an end-to-end network, controlling steering from single camera for lane following on empty roads. \cite{hubschneider2017adding} adds conditional navigation to this same approach.
\cite{codevilla2018end} learn longitudinal and lateral control via conditional imitation learning, following route commands on a remote control car in a static environment.
\cite{amini2018variational} develop a steering-only system that learns to navigate a road network using multiple cameras and a 2D map for localisation, however it uses heavily cropped images, and is dependent on localisation and route map generation.

The CARLA simulator~\cite{dosovitskiy2017carla} has enabled significant work on learning to drive.
One example is the work of \cite{codevilla_2019_bc}, which established a new behaviour cloning benchmark for driving in simulation.
However, simulation cannot capture real-world complexities, and achieving high performance in simulation is significantly simpler due to a state space with less entropy, and the ability to generate near-infinite data.
Several approaches have transferred policies from simulation to the real world.
\cite{mueller_driving_2018} use semantic segmentation masks as input and waypoints as output.
\cite{bewley2018learning} learn a control latent space that allows domain transfer between simulation and real world.
Sim-to-real approaches are promising, but policy transfer is not always effective and robust performance requires careful domain randomisation~\cite{andrychowicz2018learning, tobin2018domain, peng2018sim}.

Rather than learning a driving policy from egocentric camera images, \cite{bansal_chauffeurnet:_2018} and \cite{Zeng_2019_CVPR} present approaches to using imitation learning based on a bird's eye view (BEV) of a scene (using a fused scene representation and LiDAR data respectively, neither being fully end-to-end due to the need for prior perception and/or mapping). However, both approaches require additional infrastructure in the form of perception, sensor fusion, and high-definition (HD) mapping: in preference to this, we advocate for learning from raw sensor (camera) inputs, rather than intermediate representations.

Recently, model-based reinforcement learning (RL) for  learning driving from simulated LiDAR data by \cite{rhinehart_2018}, but it has yet to be evaluated in real urban environments. Approaches with low dimensional data have shown promising results in off-road track driving \cite{williams2017information}.
Model-free RL has also been studied for real-world rural lane following \cite{kendall2018learning}.

Our work fits in the end-to-end imitation learning literature and takes inspiration from the much of the work here. We extend the capabilities of a driving policy beyond what has been demonstrated to date: learning full lateral and longitudinal control on a vehicle with conditional route following and behaviour with traffic.
We develop new methods to deploy our system to real world environments. Specifically, we focus on urban streets, with routes not demonstrated during training, evaluated over 35km of urban driving.

\section{Method}
\label{sec:method}
%






\begin{figure}
    \centering
    \includegraphics[width=1.0\columnwidth]{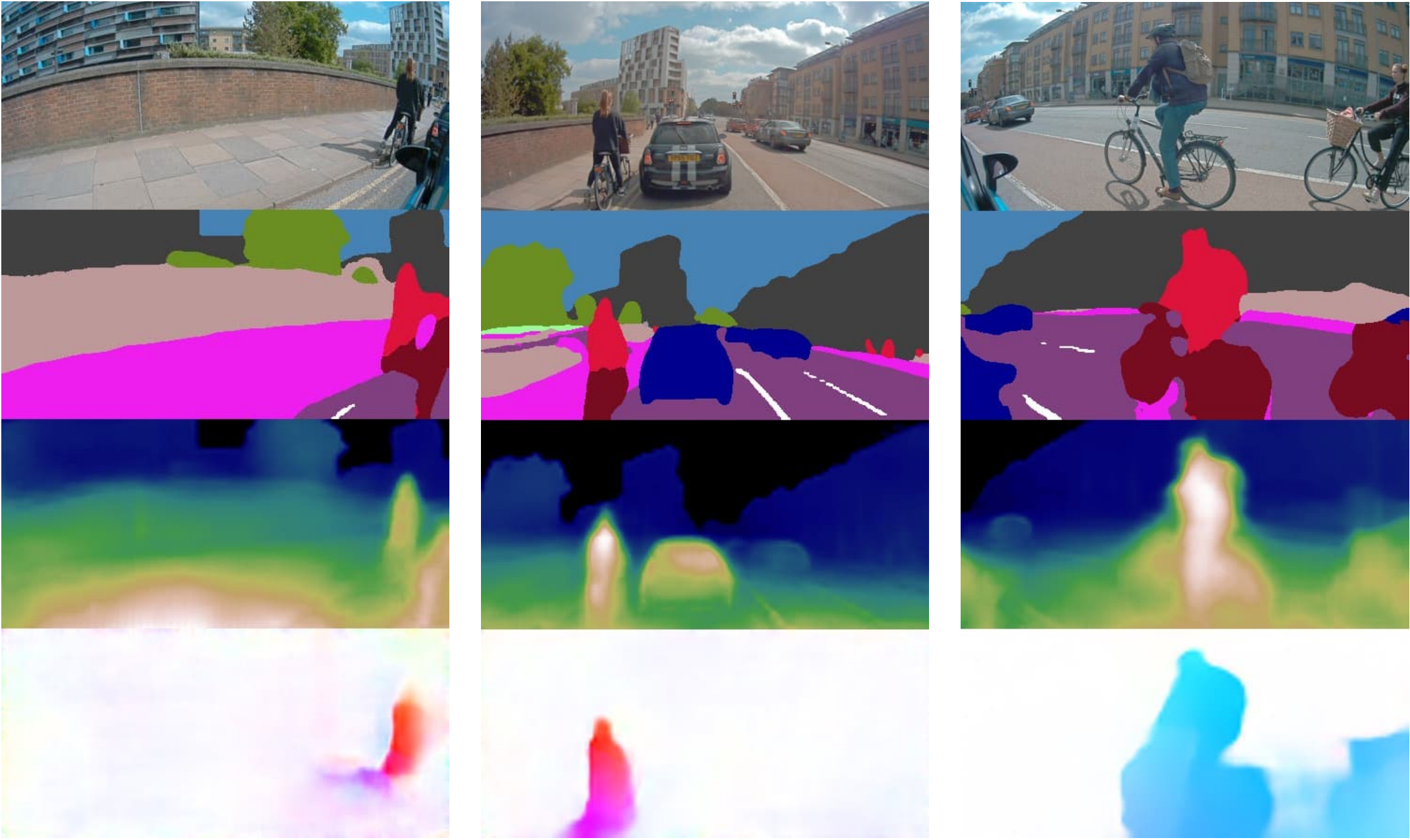}
    \caption{An example of perception from the intermediate learned features of the three forward-facing cameras on our vehicle in a typical urban scene (forward-left, forward, and forward-right). From top to bottom: RGB input, segmentation, monocular depth, and optical flow.}
    \label{fig:perception_example}
\end{figure}

We begin with a brief introduction of imitation learning (IL) and conditional IL, followed by a discussion of our method and model architecture. 

Consider a dataset of observation-action pairs $\gD\{\langle \rvo_i, \rva_i \rangle\}_{i=1}^N$, collected from expert demonstrations.
The goal of IL is to learn a policy $\pi_\theta(\rvo_t): \gO \rightarrow \gA$ that maps observations $\rvo_t$ to actions $\rva_t$ at every time step and is able to imitate the expert.
The parameters $\theta$ of the policy are optimized by minimizing a distance $\gL$ between the predicted action and the expert action:
\begin{equation}
    \min_{\theta} \sum_{i} \gL \big(\pi_\theta(\rvo_i), \rva_i \big)
\end{equation}
Conditional IL (CIL) \cite{codevilla2018end} seeks to additionally condition the policy on some high-level command $\rvc$ that can convey the user intent at test time.
Such a command can serve to disambiguate multi-modal behaviour: for example, when a car approaches an intersection, the camera input is not sufficient to predict whether the car should turn left or right, or go straight.
Providing a route command $\rvc$ helps resolving the ambiguity.
The CIL objective can be written as:
\begin{equation}
    \min_{\theta} \sum_{i} \gL \big(\pi_\theta(\rvo_i, \rvc_i), \rva_i \big)
\end{equation}

\subsection{Model Architecture}
\label{sec:model-arch}

Our method learns directly from images and outputs a local motion plan for speed and steering.
The driving policy is a fully end-to-end neural network, but we conceptually structure it with three separate components: \textit{perception}, \textit{sensor fusion} and \textit{control}.
We use a combination of pretrained and trainable layers to learn a robust driving policy.
Figure~\ref{fig:mcn-architecture} outlines the architectural details discussed below.

\subsubsection{Perception}
\label{subsubsec:perception}
The perception component of our system consists of a deep encoder-decoder similar to multitask segmentation and monocular depth~\cite{Kendall_2018_CVPR}.
It receives an image as observation and is trained to reconstruct RGB, depth and segmentation.
For driving, we use the encoded features, which contain compressed information about the appearance, the semantics, and distance estimation.
In principle, such representations could be learned simultaneously with the driving policy, for example, through distillation. Alternatively, the labelled control data could be labelled for perceptual tasks.
However, to improve data efficiency and robustness when learning control, we pretrain the perception network on several large, heterogeneous, research vision datasets~\cite{yu2018bdd100k, MVD2017, cordts2016cityscapes, Deeplab, Geiger2013IJRR}.

The perception architecture above does not have any temporal information about the scene.
One solution is to use concatenated representations of the past $n$ frames, but this did not perform well in our experiments.
Instead, we use intermediate features from an optical flow model similar to \cite{sun2018pwc}, and concatenate this to form features rich in semantics, geometry, and motion information.

Apart from the single forward-facing camera, in some of our experiments we additionally use a left-facing and a right-facing view.
In this case, the perception component treats each of these frames independently.
Figure~\ref{fig:perception_example} provides an example of the perception output on input images.

\subsubsection{Sensor Fusion}
\label{subsubsec:sf}

The purpose of the sensor fusion component is to aggregate information from the different sensors and process them into a single representation of the scene, useful for driving.


\textbf{Singleview}. In the case of a single camera, the model is composed of a simple feedforward architecture of convolutional and fully-connected layers.

\textbf{Multiview}.
Based on single-frame observations from 3 cameras: front, left and right (Figure~\ref{fig:mcn-architecture}). 
Driving with a single view means the model must predict an action without full observability of the necessary state (e.g., losing sight of the adjacent kerb while approaching a junction). Enhancing observability with multiple views should improve behaviour in these situations.
However, we found that naively incorporating all camera features led to over-reliance on spurious information in the side views.
Thus, we occasionally corrupt the side view encoder outputs during training, using the augmentation procedure described below for flow. We also apply self-attention \cite{zhang2018sagan} on the perception output of each camera, allowing the network to model long-range dependencies across the entire image and focus on the parts of the image that are important for driving.

\textbf{Multiview and Flow}. Additionally to the Multiview architecture above, optical flow is computed only for the front camera using the last two consecutive frames. The flow features are concatenated to the perception features for the front camera and then processed by sensor fusion.

Access to more information, such as motion, can actually lead to worse performance. As discussed by \cite{codevilla_2019_bc, causalconf}, imitation learning in particular can suffer from \textit{causal confusion}: unless an explicit causal model is maintained, spurious correlations cannot be distinguished from true causes in the demonstrations.
For example, inputting the current speed to the policy causes it to learn a trivial identity mapping, making the car unable to start from a static position.
While \cite{causalconf} propose a way to learn a causal graph, it assumes access to a disentangled representation and its complexity grows exponentially with the number of features.

Instead, we propose a method, which cannot determine the causal structure, but can  overcome the causal confusion problem and scale to large number of features.
Given an input $\rvx$, we add random noise, e.g. Gaussian, and apply dropout \cite{dropout} on the noise with probability $0.5$ for the full duration of training.
This allows the model to use the true value of $\rvx$ when available and breaks the correlation the rest of the time, forcing the model to focus on other features to determine the correct output.
Applying this approach on the flow features during training allowed the model to use explicit motion information without learning the trivial solution of an identity mapping for speed and steering.

\subsubsection{Control}
\label{subsubsec:control}

The control module consists of several fully-connected layers that process the representation computed by sensor fusion.
At this stage, we also input a driving route command $\rvc_t$, corresponding to one of \textit{go-straight}, \textit{turn-left} or \textit{turn-right}, as a one-hot-encoded vector.
We found that inputting the command multiple times at different stages of the network improves robustness of the model.

We selected this encoding of a route to ground this approach with those demonstrated in CARLA \cite{codevilla2018end, codevilla_2019_bc, mueller_driving_2018, rhinehart2018deep}. In practice, this limits real world testing to grid-like intersections. We use this encoding for the approach outlined here, but we favour a richer learned route embedding similar to \cite{amini2018variational, hecker_end--end_2018}.

The output of the control module consists of a simple parameterised motion plan for lateral and longitudinal control.
In particular, we assume that the vehicle motion is locally linear and we output a line, parameterised by the current prediction $\rvy_t$ and a slope $\rvm_t$, for both speed and steering.
During training, we minimise the mean squared error between the expert actions, taken over $N$ timesteps into the future, and the corresponding actions predicted by the network motion plan:
\begin{equation}
    \sum_{n=0}^{N-1} \gamma^{\Delta_n} \left(\rvy_t + \rvm_t \Delta_n - \rva_{t+n} \right)^2
    \label{eq:loss}
\end{equation}
where $\gamma$ is a future discount factor, $\Delta_n$ is the time difference between steps $t$ and $t+n$, $\rvy_t$ and $\rvm_t$ are the outputs of the network for $\rvo_t$, and $\rva_{t+n}$ is the expert speed and steering control at timestep $t+n$. Predicting such trend into the future provides for smoother vehicle motion and demonstrated better performance both in closed loop and open loop testing.

\subsection{Data}
\label{sec:data}
Driving data is inherently heavily imbalanced, where most of the captured data will be driving near-straight in the middle of a lane, as shown in Figure \ref{fig:data-dist}.
To mitigate this, during training we sample data uniformly across lateral and longitudinal control dimensions.
We found that this avoids the need for data augmentation~\cite{bansal_chauffeurnet:_2018} or synthesis~\cite{bojarski_end_2016}, though these approaches can be helpful.

We split the data into $k\ge 1$ bins by steering value, defining $x_0, x_k$ as the minimal and maximal steering values and the bin edges $x_1, \ldots, x_{k-1}$.
We define the edges such that the product of the number of data in each bin and the width of that bin (i.e. $x_j - x_{j-1}$) are equal across all bins, finding these edges using gradient descent.
Having defined the edges, we then assign weights to data:
\begin{equation}
    w_j = \frac{x_j - x_{j-1}}{N_j}\cdot\frac{W}{x_k - x_0},
\end{equation}
where $w_j$ denotes the weight of data in the $j$-th bin, $N_j$ --- the number of data in that bin and $W$ --- the total weight of the dataset (equal to the number of data).
The total weight of each bin is proportional to its width as opposed to the number of data.
The second factor is for normalisation: it ensures that the total dataset weight is $W$. We recursively apply this to balance each bin w.r.t. speed.


\section{Experiments}
\label{sec:experiments}
%

%
%
%
%
%
%

The purpose of our experiments is to evaluate the impact of the following features on the driving performance of the policy:
\begin{itemize}
  \item Using computer vision to learn explicit visual representations of the scene, in contrast to learning end-to-end strictly from the control loss,
  \item Improving the learned latent representation by providing wider state observability through multiple camera views, and adding temporal information via optical flow,
  \item Understanding influence of training data diversity on driving performance.
\end{itemize}

\label{sec:models}
To assess these questions, we train the following models:
\begin{itemize}
    \item SV: Single view model.
    Based on the network described in Section~\ref{sec:model-arch}, using only the forward facing camera.
    The parameters of the perception module are pretrained and frozen while training the sensor fusion and the control modules.
    A local linear motion plan for speed and steering is predicted over $1$s in the future.
    This model serves as our baseline.
  \item MV: Multiview model. As SV, but camera observations from forward, left, and right views are integrated by the sensor fusion module as described in \ref{subsubsec:sf}.
  \item MVF: Multiview and Flow model. As MV, but with additional optical flow information.
  \item SV75: As SV, but using only the first 75\% of the training data. The latter 25\% of the data was discarded, degrading the diversity while retaining comparable temporal correlation. 
  \item SV50: As SV75, but using only 50\% of the training data (discarding the most recent 50\%).
  \item SV25: As SV75, but using only 25\% of the training data (discarding the most recent 75\%).
  \item SVE2E: Same as SV, but trained fully end-to-end with randomly initialised parameters.
  \item SVE2EFT: E2E fine-tuned perception. SV model with a second E2E fine-tuning stage.
  \item SVE2EPT: Same as SVE2E, but perception parameters pretrained as in SV.
\end{itemize}

We evaluate all of the above models in closed loop on real-world urban European streets using the procedure described in Section \ref{subsec:procedure} and the metrics discussed in \ref{subsec:metrics}.
We note that we also evaluated the baseline model SV in simulation with an in-house simulator: its performance was significantly better compared to the real world, highlighting the simplification of the problem by simulation.

\subsection{Procedure}
\label{subsec:procedure}

\textbf{Data Collection}. 
We collected data from human demonstrations over the span of 6 months for a total of 30 driving hours in a densely populated, urban environment representative of most European cities.
The drivers involved were not given any explicit instructions and were free to choose random routes within the city.
Collected data includes front, left and right cameras as well as measurements of the speed and steering controls from the driver.
Images were captured from rolling shutter cameras, synchronised within $5$ms with a frame rate of 15Hz.
Scalar values (e.g., speed) were received at 100Hz.
We used the front camera as the primary clock, associating this with the latest received values for the rest of the sensors. Route annotations were added during postprocessing of the data, based on the route the driver took.
Table \ref{tab:data} and Figure \ref{fig:data-dist} summarise the training and test datasets.

\begin{table}[hbtp]
	\begin{center}
	\begin{tabular}{l|c c c c}
	Dataset & Frames & Distance (km) & Time (hours) & Drivers \\
	\hline
	Training & 735K & 613 & 27.2 & 8\\
	Test & 91K & 89 & 3.4 & 1
	\end{tabular}
	\end{center}
	\vspace{-2mm}
	\caption{Training and test driving datasets, collected over a 6-month period in a typical European city.}
	\label{tab:data}
\end{table}

\begin{figure*}[hbtp]
\centering
  \begin{subfigure}[b]{0.3\textwidth}
    \centering
    \includegraphics[height=2.5cm]{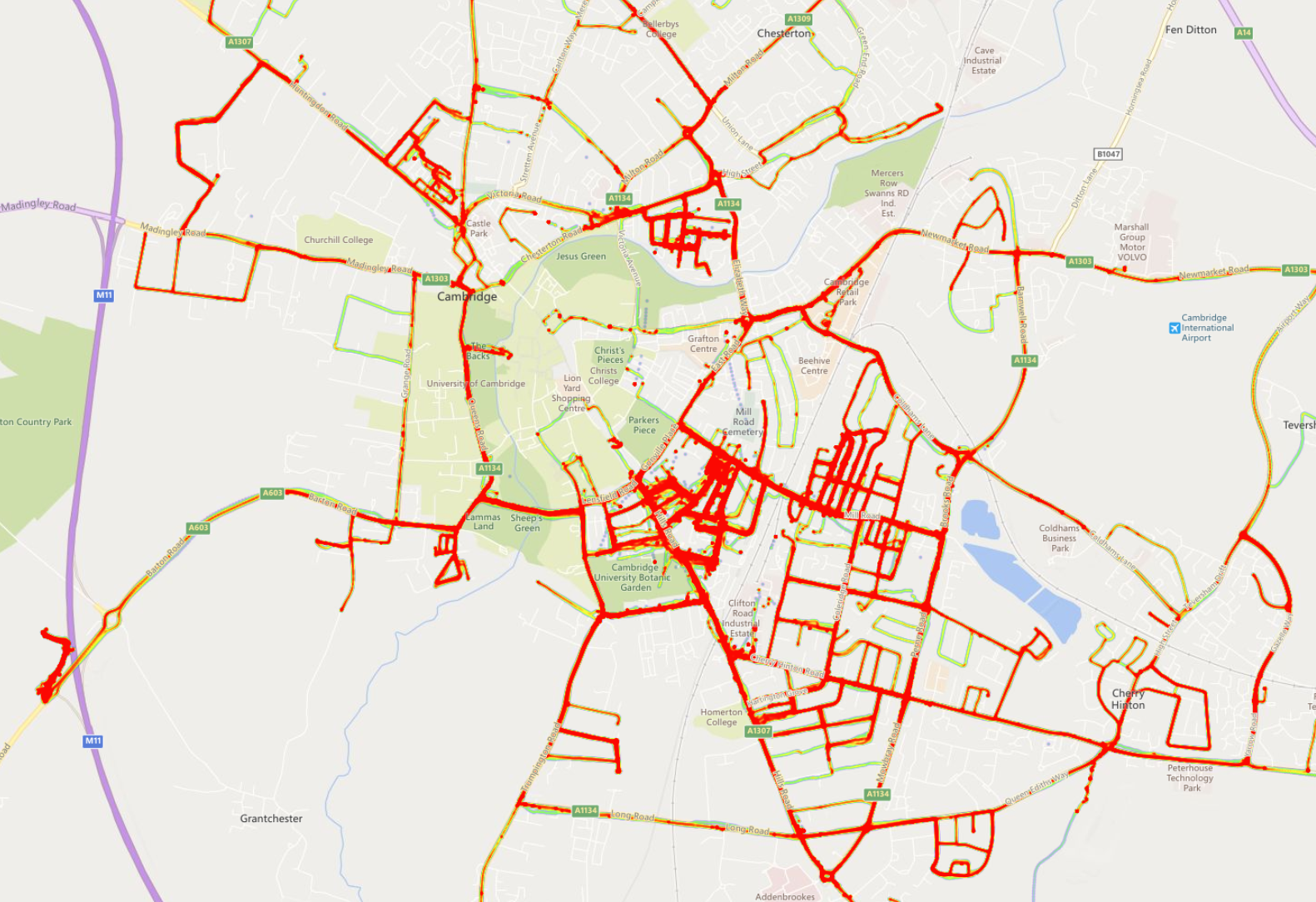}
    \caption{Training collection area}
    \label{fig:training-gps}
  \end{subfigure}
  \begin{subfigure}[b]{0.3\textwidth}
       \centering
       \includegraphics[height=2.5cm]{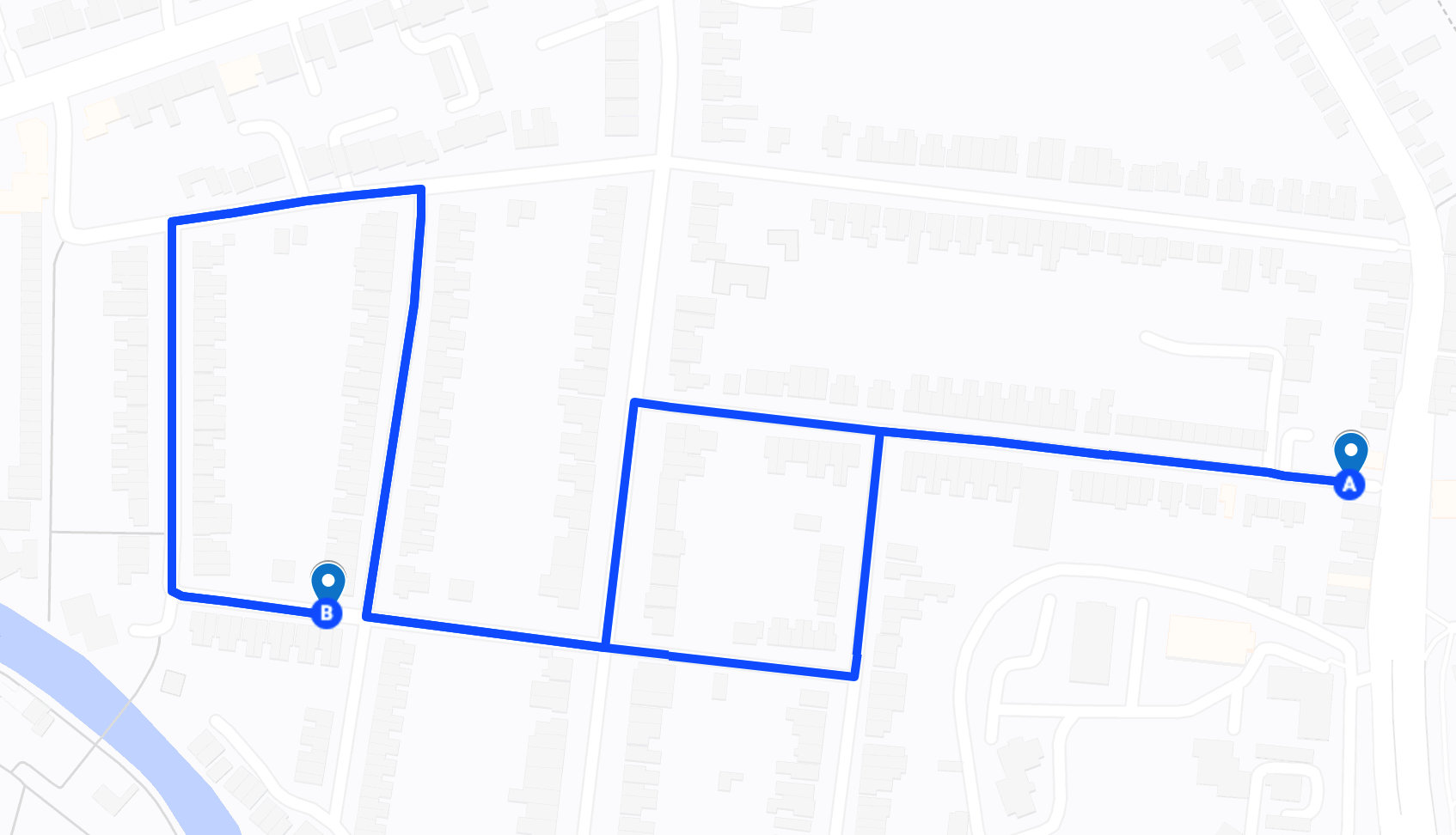}
       \caption{Test route A}
       \label{fig:chesterton}
  \end{subfigure}
  \begin{subfigure}[b]{0.3\textwidth}
       \centering
       \includegraphics[height=2.5cm]{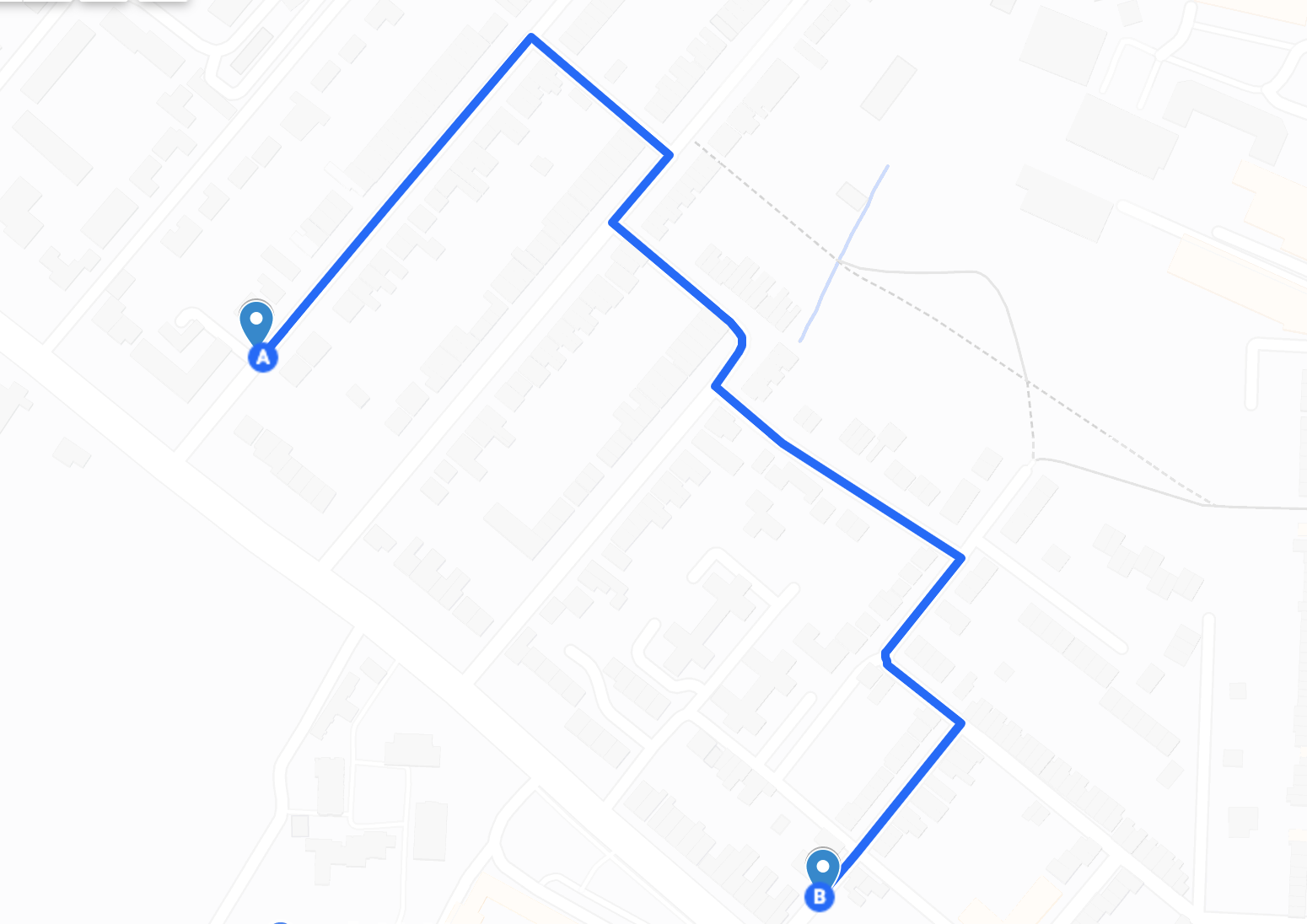}
       \caption{Test route B}
       \label{fig:oxfordrd}
  \end{subfigure}
  \caption{
  We collect exemplar driving data by driving through the European urban area in \ref{fig:training-gps}. We evaluate the learned driving policy performance in two urban routes, each approximately 1km in length with multiple intersections. Both routes are dense urban scenes which broadly fall into a Manhattan-like grid structure, with a mix of road widths. Models tested on each route are evaluated in both directions: note that route A takes a slightly different route on the return due to safety-driver visibility at an intersection. The driving environment is comparable to the scene shown in Figure \ref{fig:perception_example}.}
  \label{fig:expt-routes}
\end{figure*}

\textbf{Evaluation Procedure}
We evaluate the ability of a learned driving system to perform tasks that are fundamental to urban driving: following lanes, taking turns at intersections and interacting with other road agents.
We select two routes not present in the training data in the same city, each approximately $1$km in length and measure the intervention rates while autonomously completing the manoeuvres required in order to reach the end of the route from the start.
Each route was selected to have dense intersections, and be a largely Manhattan-like environment analogous to ~\cite{codevilla2018end} to avoid ambiguity with the route encoding.
Figure \ref{fig:expt-routes} shows the locations of these routes.
Completing a route in a busy urban environment implicitly entails navigating around other dynamic agents such as vehicles, cyclists and pedestrians, to which the model has to respond safely through actions such as stopping or giving way.

Each model was commanded to drive the route in both directions, each of which counts as an attempt.
Should a given model have five interventions within a single route attempt, the attempt was terminated early. Models with consistently poor performance were evaluated on fewer attempts.

We additionally evaluate these models interacting with another vehicle, following a pace car which periodically stopped throughout the driving route. During these tests, we measure intervention rates for stopping behind a stationary vehicle or failing to follow the pace car.

The interventions as well as their types are left to the judgment of the safety-driver, who is able to take over control whenever the model exhibits poor behaviour.
Evaluating a driving policy's performance in terms of the number and type of interventions is preferable to computing low level metrics (\eg, cross-track error to a path) as there are multiple solutions to performing the task correctly.

We note that neither of the testing routes has been seen by the model during training, though the data has some overlap with some sections.
While stopping for and following vehicles is a common occurrence in the data, the particular test vehicle used was not seen during training.

Finally, the testing environment will inevitably contain time-dependent factors beyond our control - this is expected for outdoor mobile robot trials. Examples include the position and appearance of parked and dynamic vehicles, weather conditions, presence and behaviour of other agents, however, we have controlled for these factors as much as possible by running trials with batches of models in succession. Fundamentally, this real-world constraint forces a driving system to generalise: each attempt of the same route is novel in a different way, e.g., weather, lighting, road users, static objects. 


\subsection{Training Procedure}
\label{app:train}

The network described in Sections~\ref{sec:model-arch} is trained jointly using stochastic gradient descent (SGD) using batches of size $256$ for $200k$ iterations. The initial learning rate is set to $0.01$ and decayed linearly during training, while momentum and weight decay are set to $0.9$ and $0.0001$ respectively.

In total, the network architecture used here consisted of $13-26$M trainable parameters, varying depending on the perception encoder used, and the number of cameras. 

\begin{table}[hbtp]
	\begin{center}
	\begin{tabular}{l | c c c c}
	Model & Perception & Sensor Fusion & Control & Total \\
	\hline
	SV & 10.4M & 2.2M & 705K & 13.3M \\
	MV & 10.4M & 6.5M & 442K & 17.4M \\
	MVF & 19.1M & 6.5M & 442K & 26.1M \\
	\end{tabular}
	\end{center}
	\caption{Number of trainable parameters in each conceptual group within the network architectures used here.}
	\label{tab:num-params}
\end{table}

\subsection{Metrics}
\label{subsec:metrics}

We need a principled way of identifying which of the models may perform best under closed-loop test conditions.
For this purpose we use open-loop metrics of the control policy, computed on a hold-out validation dataset.
As discussed by \cite{Codevilla_2018_ECCV} and~\cite{Liang_2018_ECCV}, the correlation between offline open-loop metrics and online closed-loop performance is weak.
We observed similar results, but we found some utility in using open-loop weighted mean absolute error for speed and steering as a proxy for real world performance (referred to in Table \ref{tab:results} as Balanced MAE). 
In particular, we used this metric on the test dataset in Table \ref{tab:data} to select models for deployment.
For the experimental results, we adopt the following closed-loop metrics:

\begin{figure}[hbtp]
  \centering
  \begin{subfigure}[b]{0.2\textwidth}
       \centering
       \includegraphics[height=2.2cm]{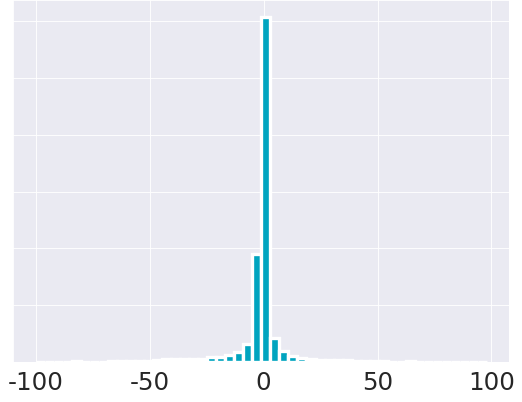}
       \caption{Steering ($\pm$ \%)}
       \label{fig:steering}
  \end{subfigure}
  \begin{subfigure}[b]{0.2\textwidth}
       \centering
       \includegraphics[height=2.2cm]{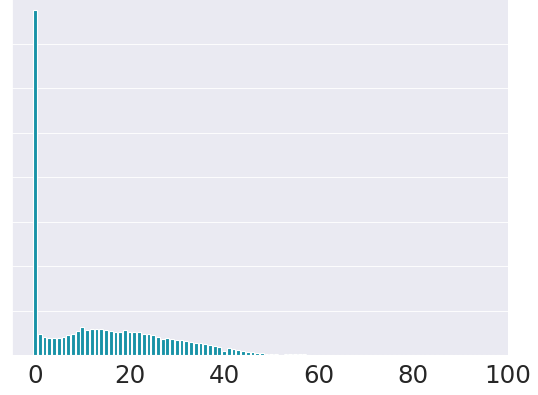}
       \caption{Speed (km/h)}
       \label{fig:speed}
  \end{subfigure}
  \begin{subfigure}[b]{0.2\textwidth}
       \centering
       \includegraphics[height=2.22cm]{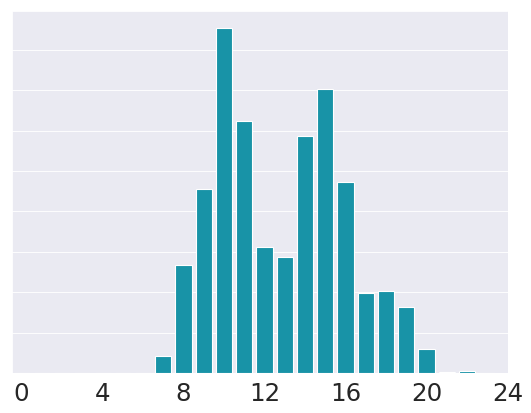}
       \caption{Time of day (hour)}
       \label{fig:timestamp}
  \end{subfigure}
  \begin{subfigure}[b]{0.2\textwidth}
       \centering
       \includegraphics[height=2.2cm]{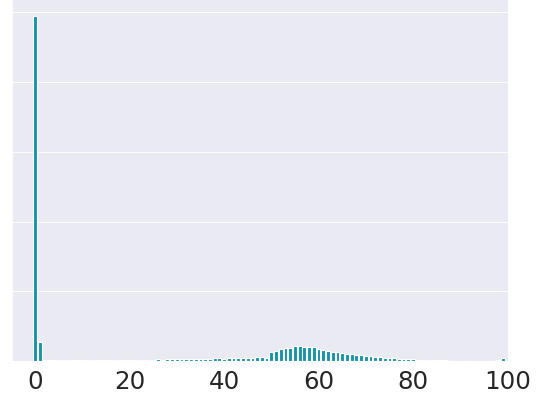}
       \caption{Throttle (\%)}
       \label{fig:throttle}
  \end{subfigure}
  \caption{We collect training data driving across a European city (see the heatmap in \ref{fig:training-gps}, showing data density). The data, as is typical, is imbalanced, with the majority driving straight (\ref{fig:steering}), and a significant portion stationary (\ref{fig:speed}).}
  \label{fig:data-dist}
\end{figure}

\begin{itemize}
  \item Intervention rate, computed as metres per intervention,
  \item Intervention rate only during lane following (a subset of the previous metric),
  \item Success rate of turning manoeuvres (left, right),
  \item Intervention rate while following a pace car, travelling at up to $15$ km/h,
  \item Success rate of stopping behind a pace car.
\end{itemize}

In addition to the rate of interventions, we also  provide their type using the following 
 categories:
\begin{itemize}
  \item \emph{Obstacle}: \eg, prevention of crashing into an obstacle such as a parked car or road works, 
  \item \emph{Road position}: \eg, wrong side of traffic or dangerous proximity to the kerb,
  \item \emph{Disobeyed navigation}: \eg, drove straight instead of turning left,
  \item \emph{Failed manoeuvre}: \eg, attempted the correct turn, but failed to complete the manoeuvre,
  \item \emph{Dynamic vehicle}: \eg, acted poorly with other traffic, such as not stopping behind a car.
\end{itemize}

Finally, while these metrics provide quantitative results, they fail to capture qualitative properties.
We therefore provide some commentary on the models' observed behaviour.

\subsection{Results}

Table \ref{tab:results} outlines the experimental results. A total of $34.4$km of driving was performed under automation, prioritising the SV, MV, and MVF models. We consider the performance in a static environment (\emph{driving manoeuvres}) independently of the performance following a pace car (\emph{traffic following}). 

\begin{table*}[hbt]
\centering
\resizebox{\textwidth}{!}{
    \begin{tabular}{lccccccccccc} 
    \toprule
      \multirow{2}*{Model} & \multicolumn{2}{c}{Attempts} & \multicolumn{2}{c}{Distance (m)} & \multicolumn{2}{c}{Intervention rate (m/int)} & \multicolumn{3}{c}{Success Rate (\%)} & \multicolumn{2}{c}{Open Loop} \\
      \cmidrule(lr){2-3} \cmidrule(lr){4-5} \cmidrule(lr){6-7} \cmidrule(lr){8-10} \cmidrule(lr){11-12}
       & A & B & Manoeuvres (LF only) & Traffic follow & Manoeuvres (LF only) & Traffic Follow & Turn: Left & Turn: Right & Stop: Car & MAE & Bal-MAE  \\ 
      \midrule
      SV & 14 & 12 & 5113 (4094) & 3158 & 222 (372) & 121 & 74\% (N=31) & 87\% (N=31) & 69\% (N=36) & 0.0715 & 0.0630 \\
      MV & 14 & 12 & 5811 (4613) & 3440 & 306 (659) & 191 & 70\% (N=23) & 86\% (N=36) & 65\% (N=40) & 0.0744 & 0.0706 \\
      MVF & 14 & 12 & 5313 (4086) & 2569 & 253 (511) & 161 & 78\% (N=32) & 82\% (N=34) & 81\% (N=36) & 0.0633 & 0.0612 \\
      \midrule
      SV75 & 2 & 2 & 2777 (2258) & - & 154 (376) & - & 57\% (N=13) & 80\% (N=17) & - & 0.0985 & 0.0993\\
      SV50 & 2 & 4 & 2624 (2145) & - & 67 (98) & - & 54\% (N=19) & 50\% (N=16) & - & 0.0995 & 0.1015 \\
      SV25 & 2 & - & 926 (746) & - & 44 (50) & - & 33\% (N=6) & 71\% (N=7) & - & 0.1081 & 0.1129 \\
      \midrule
      SVE2E & 2 & 2 & 823 (615) & - & 30 (30) & - & 42\% (N=12) & 100\% (N=5) & - & 0.1410 & 0.1365 \\
      SVE2EFT & 2 & 2 & 1591 (759) & - & 177 (379) & - & 69\% (N=16) & 90\% (N=21) & - & 0.0769 & 0.0801 \\
      SVE2EPT & 2 & 2 & 260 (177) & - & 26 (59) & - & 33\% (N=3) & 0\% (N=5) & - & 0.0966 & 0.0946 \\
      \bottomrule
    \end{tabular} 
}
\caption{
Driving performance metrics evaluated on routes A and B (see \ref{sec:models} and \ref{subsec:procedure}).
\emph{Traffic Follow} and \emph{Manoeuvres} present the metrics for following the routes with and without a pace car.
The \emph{LF only} metrics refer to lane following.
In \emph{Success Rate}, $N$ refers to the number of manoeuvre attempts.
}
\label{tab:results}
\end{table*}
\begin{figure*}[t!]
    \begin{subfigure}[b]{0.33\textwidth}
        \centering
        \includegraphics[height=4.4cm]{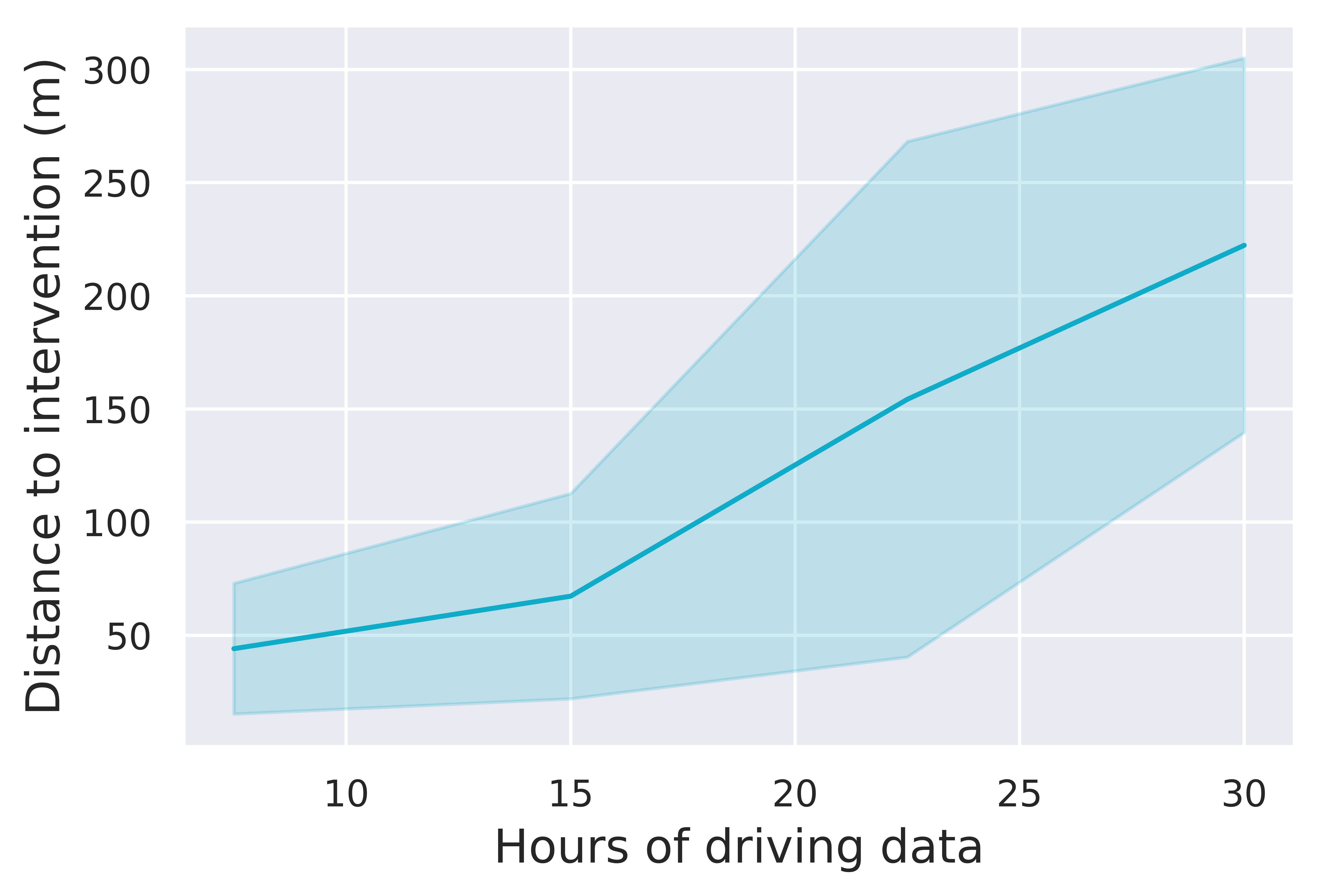}
        \caption{Performance trend with increasing data}
        \label{fig:data-quantity}
    \end{subfigure}
 	\hspace{1.4cm}
    \begin{subfigure}[b]{0.55\textwidth}
        \includegraphics[height=4.5cm]{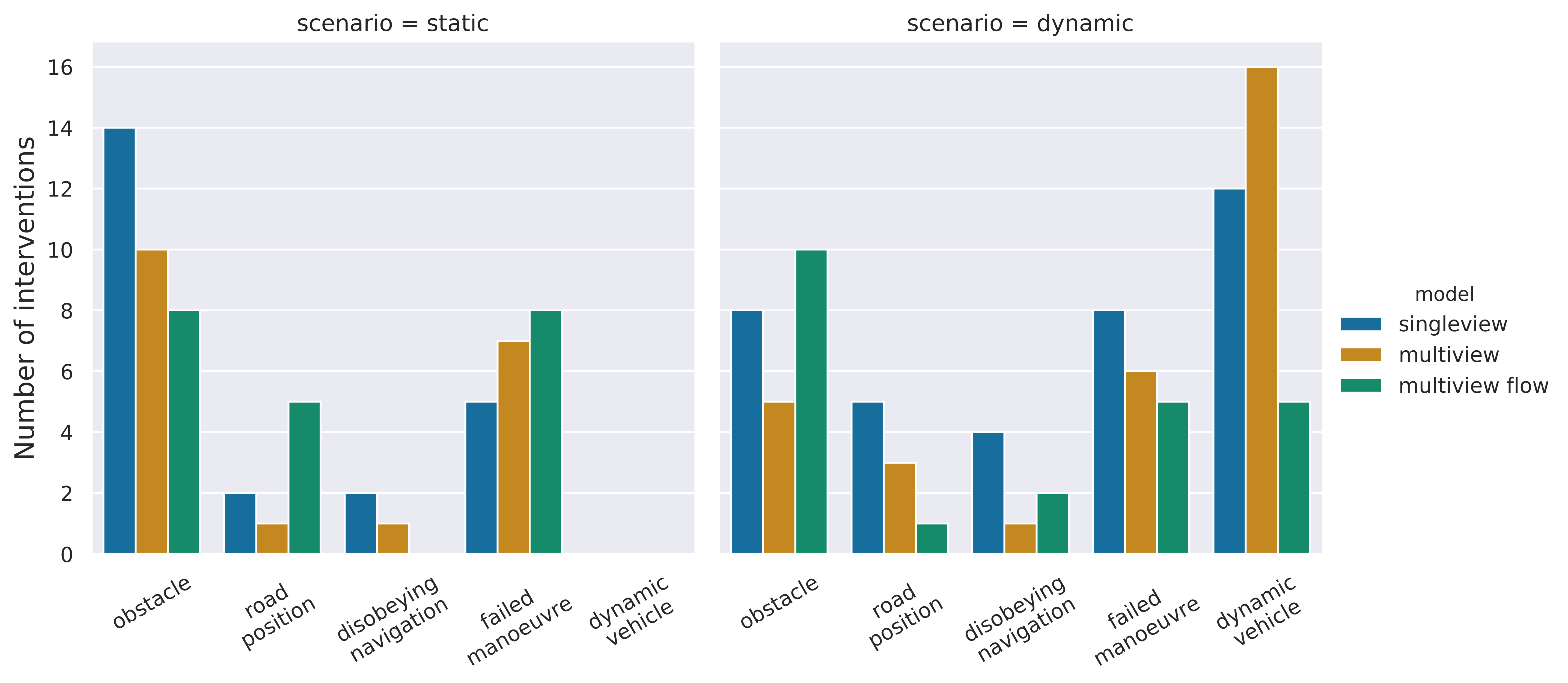}
        \caption{Interventions by type}
        \label{fig:interventions}
    \end{subfigure}
    \caption{
From the results in Table \ref{tab:results} we make a number of observations. In \ref{fig:data-quantity} we see a clear relationship between data quantity and SV model performance (showing mean, s.d.). Inspecting the interventions types in \ref{fig:interventions} we observe that the addition of optical flow improves dynamic vehicle behaviour, but potentially suffers due to a lower signal-to-noise ratio with increased dimensionality.
    }
    \label{fig:results}
\end{figure*}

\subsubsection{Representation Learning with Computer Vision}

Comparing the performance of SV to the E2E trained models (SVE2E, SVE2EFT, and SVE2EPT) suggests that learned scene understanding is a critical component for a robust representation for driving. Both fully E2E trained models (SVE2E, SVE2EPT) perform very poorly. 

We attribute this to the fact that our perception model is significantly more robust to the diversity of appearance present in real world driving images \cite{RobotCarDatasetIJRR}, in part influenced by the additional data seen during the initial perception training phase. In robotics we must consider methods which prioritise data efficiency: we will never have the luxury of infinite data. For example, here we train on 30 hours of driving, not nearly sufficient to cover the diversity of appearance in real world scenes.

We observed that the E2E trained models performed comparably to SV when operating in shaded environments, but poorly in bright sunlight. We attribute this to the appearance change between our training data and the test environment: the data was gathered in winter, three months prior to tests being conducted in spring.
Additionally, the fine-tuned model performs similarly to the baseline: on this basis we recommend using pretrained perception due to the wider benefits of  interpretability.

\subsubsection{Improving the learned state: observability and temporal information}

Firstly we consider the performance of multiview models compared to the the singleview baseline. We see an improvement in driving performance when providing the model with a more complete view of the scene. This is apparent when considering the \emph{disobeyed navigation} interventions: the SV model systematically fails to take certain turns. For example, if the side road is narrow, it tends to drive straight past, ignoring the route. Equally, the model commonly cuts the near-side kerb while turning. Both these failure modes are attributable to the fact that the forward facing camera has no visibility of this part of the scene. Multiview models tend not to exhibit these failure modes. Qualitatively, we found the MV models to provide a much larger buffer around the kerb, however both models would exhibit the same failure mode of poor positioning behind parked cars around turns without space to correct. We attribute this failure mode to the low perception resolution and rather simple action representation used in this work.

Secondly, we consider the benefit of augmenting the learned representation with temporal state, describing the motion present in the scene with optical flow. MVF performs slightly worse than MV in these metrics. We attribute this to a lower signal-to-noise ratio given the increased dimensionality, though we consider these to be largely comparable. Where this motion information clearly has benefit is behaviour with other vehicles (stopping): the model with motion information responds to other agents more robustly, demonstrated by a success rate increase from 65\% to 81\%.

\subsubsection{Influence of data quantity and diversity on performance}

The learned driving policies presented here need significant further work to be comparable to human driving. We consider the influence of data quantity on SV model performance, observing a direct correlation in Figure \ref{fig:data-quantity}. Removing a quarter of the data notably degrades performance, and models trained with less data are almost undriveable. 
We attribute this to a reduction in data diversity. As with the E2E trained models, we found that the reduced data models were able to perform manoeuvres in certain circumstances. Qualitatively, all models tested perform notably better during environmental conditions closer to the collection period (concluded three months prior to these experiments). We observed that, in the best case, these models drove nearly $2$km without intervention. While we do not know the upper limit on performance for the effect of data, it is clear that more data than our training set --- and more diversity --- is likely to improve performance.

\section{Conclusions} 
\label{sec:conclusion}
%
%


We present a conditional imitation learning approach which combines both lateral and longitudinal control on a real vehicle.
To the best of our knowledge, we are the first to learn a policy that can drive on a full-size vehicle with full vehicle actuation control in complex real-world urban scenarios with simple traffic.
Our findings indicate that learning intermediate representations using computer vision yields models which significantly outperform fully end-to-end trained models.
Furthermore, we highlight the importance of data diversity and addressing dataset bias.
Reduced state observability models (i.e., single view) can perform some manoeuvres, but have systematic failure modes.
Increased observability through the addition of multiple camera views helps, but requires dealing with causal confusion.

The method presented here has a number of fundamental limitations which we believe are essential for achieving (and exceeding) human-level driving.
For example, it does not have access to long-term dependencies and cannot `reason' about the road scene.
This method also lacks a predictive long-term planning model, important for safe interaction with occluded dynamic agents.
There are many promising directions for future work. The time and safety measures required to run a policy in closed loop remains a major constraint. 
Therefore, being able to robustly evaluate a policy offline and quantify its performance is important area for research.
Including the ability to learn from corrective interventions (e.g.,  \cite{dagger, kendall2018learning}) is another vital direction for future work, and for generalising to real-world complexities across different cities.

\section*{Acknowledgments}
We thank our colleagues for their support for this work: Andrew Jones, Nikhil Mohan, Alex Bewley, Edward Liu, John-Mark Allen, Julia Gomes, and Yani Ioannou. 

\bibliographystyle{IEEEtran}
\bibliography{bsd_references}

\begin{thebibliography}{10}
\providecommand{\url}[1]{#1}
\csname url@samestyle\endcsname
\providecommand{\newblock}{\relax}
\providecommand{\bibinfo}[2]{#2}
\providecommand{\BIBentrySTDinterwordspacing}{\spaceskip=0pt\relax}
\providecommand{\BIBentryALTinterwordstretchfactor}{4}
\providecommand{\BIBentryALTinterwordspacing}{\spaceskip=\fontdimen2\font plus
\BIBentryALTinterwordstretchfactor\fontdimen3\font minus
  \fontdimen4\font\relax}
\providecommand{\BIBforeignlanguage}[2]{{%
\expandafter\ifx\csname l@#1\endcsname\relax
\typeout{** WARNING: IEEEtran.bst: No hyphenation pattern has been}%
\typeout{** loaded for the language `#1'. Using the pattern for}%
\typeout{** the default language instead.}%
\else
\language=\csname l@#1\endcsname
\fi
#2}}
\providecommand{\BIBdecl}{\relax}
\BIBdecl

\bibitem{Thrun:2006:SRW:1210475.1210482}
\BIBentryALTinterwordspacing
S.~Thrun, M.~Montemerlo, H.~Dahlkamp, D.~Stavens, A.~Aron, J.~Diebel, P.~Fong,
  J.~Gale, M.~Halpenny, G.~Hoffmann, K.~Lau, C.~Oakley, M.~Palatucci, V.~Pratt,
  P.~Stang, S.~Strohband, C.~Dupont, L.-E. Jendrossek, C.~Koelen, C.~Markey,
  C.~Rummel, J.~van Niekerk, E.~Jensen, P.~Alessandrini, G.~Bradski, B.~Davies,
  S.~Ettinger, A.~Kaehler, A.~Nefian, and P.~Mahoney, ``Stanley: The robot that
  won the darpa grand challenge: Research articles,'' \emph{J. Robot. Syst.},
  vol.~23, no.~9, pp. 661--692, Sep. 2006. [Online]. Available:
  \url{http://dx.doi.org/10.1002/rob.v23:9}
\BIBentrySTDinterwordspacing

\bibitem{dickmanns2002development}
E.~D. Dickmanns, ``The development of machine vision for road vehicles in the
  last decade,'' in \emph{Intelligent Vehicle Symposium, 2002. IEEE},
  vol.~1.\hskip 1em plus 0.5em minus 0.4em\relax IEEE, 2002, pp. 268--281.

\bibitem{leonard2008perception}
J.~Leonard, J.~How, S.~Teller, M.~Berger, S.~Campbell, G.~Fiore, L.~Fletcher,
  E.~Frazzoli, A.~Huang, S.~Karaman \emph{et~al.}, ``A perception-driven
  autonomous urban vehicle,'' \emph{Journal of Field Robotics}, vol.~25,
  no.~10, pp. 727--774, 2008.

\bibitem{bojarski_end_2016}
\BIBentryALTinterwordspacing
M.~Bojarski, D.~D. Testa, D.~Dworakowski, B.~Firner, B.~Flepp, P.~Goyal, L.~D.
  Jackel, M.~Monfort, U.~Muller, J.~Zhang, X.~Zhang, J.~Zhao, and K.~Zieba,
  ``End to {End} {Learning} for {Self}-{Driving} {Cars},'' \emph{CoRR}, vol.
  abs/1604.07316, 2016. [Online]. Available:
  \url{http://arxiv.org/abs/1604.07316}
\BIBentrySTDinterwordspacing

\bibitem{pomerleau1989alvinn}
D.~A. Pomerleau, ``Alvinn: An autonomous land vehicle in a neural network,'' in
  \emph{Advances in neural information processing systems}, 1989, pp. 305--313.

\bibitem{amini2018variational}
A.~Amini, G.~Rosman, S.~Karaman, and D.~Rus, ``Variational end-to-end
  navigation and localization,'' in \emph{2019 IEEE International Conference on
  Robotics and Automation (ICRA)}.\hskip 1em plus 0.5em minus 0.4em\relax IEEE,
  2019.

\bibitem{codevilla2018end}
F.~Codevilla, M.~Mueller, A.~L{\'o}pez, V.~Koltun, and A.~Dosovitskiy,
  ``End-to-end driving via conditional imitation learning,'' in \emph{2018 IEEE
  International Conference on Robotics and Automation (ICRA)}.\hskip 1em plus
  0.5em minus 0.4em\relax IEEE, 2018, pp. 1--9.

\bibitem{bojarski2017explaining}
M.~Bojarski, P.~Yeres, A.~Choromanska, K.~Choromanski, B.~Firner, L.~Jackel,
  and U.~Muller, ``Explaining how a deep neural network trained with end-to-end
  learning steers a car,'' \emph{arXiv preprint arXiv:1704.07911}, 2017.

\bibitem{hubschneider2017adding}
C.~Hubschneider, A.~Bauer, M.~Weber, and J.~M. Z{\"o}llner, ``Adding navigation
  to the equation: Turning decisions for end-to-end vehicle control,'' in
  \emph{2017 IEEE 20th International Conference on Intelligent Transportation
  Systems (ITSC)}.\hskip 1em plus 0.5em minus 0.4em\relax IEEE, 2017, pp. 1--8.

\bibitem{dosovitskiy2017carla}
A.~Dosovitskiy, G.~Ros, F.~Codevilla, A.~Lopez, and V.~Koltun, ``Carla: An open
  urban driving simulator,'' in \emph{Conference on Robot Learning}, 2017, pp.
  1--16.

\bibitem{codevilla_2019_bc}
F.~Codevilla, E.~Santana, A.~M. L{\'{o}}pez, and A.~Gaidon, ``Exploring the
  limitations of behavior cloning for autonomous driving,'' \emph{CoRR}, vol.
  abs/1904.08980, 2019.

\bibitem{mueller_driving_2018}
M.~Mueller, A.~Dosovitskiy, B.~Ghanem, and V.~Koltun, ``Driving {Policy}
  {Transfer} via {Modularity} and {Abstraction},'' in \emph{Proceedings of
  {The} 2nd {Conference} on {Robot} {Learning}}, vol.~87, 2018, pp. 1--15.

\bibitem{bewley2018learning}
A.~Bewley, J.~Rigley, Y.~Liu, J.~Hawke, R.~Shen, V.-D. Lam, and A.~Kendall,
  ``Learning to drive from simulation without real world labels,'' in
  \emph{IEEE International Conference on Robotics and Automation (ICRA)}, 2018.

\bibitem{andrychowicz2018learning}
M.~Andrychowicz, B.~Baker, M.~Chociej, R.~Jozefowicz, B.~McGrew, J.~Pachocki,
  A.~Petron, M.~Plappert, G.~Powell, A.~Ray \emph{et~al.}, ``Learning dexterous
  in-hand manipulation,'' \emph{arXiv preprint arXiv:1808.00177}, 2018.

\bibitem{tobin2018domain}
J.~Tobin, L.~Biewald, R.~Duan, M.~Andrychowicz, A.~Handa, V.~Kumar, B.~McGrew,
  A.~Ray, J.~Schneider, P.~Welinder \emph{et~al.}, ``Domain randomization and
  generative models for robotic grasping,'' in \emph{2018 IEEE/RSJ
  International Conference on Intelligent Robots and Systems (IROS)}.\hskip 1em
  plus 0.5em minus 0.4em\relax IEEE, 2018, pp. 3482--3489.

\bibitem{peng2018sim}
X.~B. Peng, M.~Andrychowicz, W.~Zaremba, and P.~Abbeel, ``Sim-to-real transfer
  of robotic control with dynamics randomization,'' in \emph{2018 IEEE
  International Conference on Robotics and Automation (ICRA)}.\hskip 1em plus
  0.5em minus 0.4em\relax IEEE, 2018, pp. 1--8.

\bibitem{bansal_chauffeurnet:_2018}
M.~Bansal, A.~Krizhevsky, and A.~Ogale, ``Chauffeurnet: Learning to drive by
  imitating the best and synthesizing the worst,'' in \emph{Proceedings of
  Robotics: Science and Systems}, FreiburgimBreisgau, Germany, June 2019.

\bibitem{Zeng_2019_CVPR}
W.~Zeng, W.~Luo, S.~Suo, A.~Sadat, B.~Yang, S.~Casas, and R.~Urtasun,
  ``End-to-end interpretable neural motion planner,'' in \emph{The IEEE
  Conference on Computer Vision and Pattern Recognition (CVPR)}, June 2019.

\bibitem{rhinehart_2018}
N.~Rhinehart, R.~McAllister, and S.~Levine, ``Deep imitative models for
  flexible inference, planning, and control,'' \emph{CoRR}, vol.
  abs/1810.06544, 2018.

\bibitem{williams2017information}
G.~Williams, N.~Wagener, B.~Goldfain, P.~Drews, J.~M. Rehg, B.~Boots, and E.~A.
  Theodorou, ``Information theoretic mpc for model-based reinforcement
  learning,'' in \emph{IEEE International Conference on Robotics and Automation
  (ICRA)}, 2017, pp. 1714--1721.

\bibitem{kendall2018learning}
A.~Kendall, J.~Hawke, D.~Janz, P.~Mazur, D.~Reda, J.-M. Allen, V.-D. Lam,
  A.~Bewley, and A.~Shah, ``Learning to drive in a day,'' in \emph{2018 IEEE
  International Conference on Robotics and Automation (ICRA)}, 2018.

\bibitem{Kendall_2018_CVPR}
A.~Kendall, Y.~Gal, and R.~Cipolla, ``Multi-task learning using uncertainty to
  weigh losses for scene geometry and semantics,'' in \emph{The IEEE Conference
  on Computer Vision and Pattern Recognition (CVPR)}, June 2018.

\bibitem{yu2018bdd100k}
F.~Yu, W.~Xian, Y.~Chen, F.~Liu, M.~Liao, V.~Madhavan, and T.~Darrell,
  ``Bdd100k: A diverse driving video database with scalable annotation
  tooling,'' \emph{arXiv preprint arXiv:1805.04687}, 2018.

\bibitem{MVD2017}
\BIBentryALTinterwordspacing
G.~Neuhold, T.~Ollmann, S.~Rota~Bul\`o, and P.~Kontschieder, ``The mapillary
  vistas dataset for semantic understanding of street scenes,'' in
  \emph{International Conference on Computer Vision (ICCV)}, 2017. [Online].
  Available: \url{https://www.mapillary.com/dataset/vistas}
\BIBentrySTDinterwordspacing

\bibitem{cordts2016cityscapes}
M.~Cordts, M.~Omran, S.~Ramos, T.~Rehfeld, M.~Enzweiler, R.~Benenson,
  U.~Franke, S.~Roth, and B.~Schiele, ``The cityscapes dataset for semantic
  urban scene understanding,'' in \emph{Proceedings of the IEEE conference on
  computer vision and pattern recognition}, 2016, pp. 3213--3223.

\bibitem{Deeplab}
L.~{Chen}, G.~{Papandreou}, I.~{Kokkinos}, K.~{Murphy}, and A.~L. {Yuille},
  ``Deeplab: Semantic image segmentation with deep convolutional nets, atrous
  convolution, and fully connected crfs,'' \emph{IEEE Transactions on Pattern
  Analysis and Machine Intelligence}, vol.~40, no.~4, pp. 834--848, April 2018.

\bibitem{Geiger2013IJRR}
A.~Geiger, P.~Lenz, C.~Stiller, and R.~Urtasun, ``Vision meets robotics: The
  kitti dataset,'' \emph{International Journal of Robotics Research (IJRR)},
  2013.

\bibitem{sun2018pwc}
D.~Sun, X.~Yang, M.-Y. Liu, and J.~Kautz, ``Pwc-net: Cnns for optical flow
  using pyramid, warping, and cost volume,'' in \emph{Proceedings of the IEEE
  Conference on Computer Vision and Pattern Recognition}, 2018, pp. 8934--8943.

\bibitem{zhang2018sagan}
H.~Zhang, I.~Goodfellow, D.~Metaxas, and A.~Odena, ``Self-attention generative
  adversarial networks,'' \emph{arXiv preprint arXiv:1805.08318}, 2018.

\bibitem{causalconf}
\BIBentryALTinterwordspacing
P.~de~Haan, D.~Jayaraman, and S.~Levine, ``Causal confusion in imitation
  learning,'' \emph{CoRR}, vol. abs/1905.11979, 2019. [Online]. Available:
  \url{http://arxiv.org/abs/1905.11979}
\BIBentrySTDinterwordspacing

\bibitem{dropout}
N.~Srivastava, G.~Hinton, A.~Krizhevsky, I.~Sutskever, and R.~Salakhutdinov,
  ``Dropout: A simple way to prevent neural networks from overfitting,''
  \emph{Journal of Machine Learning Research}, vol.~15, pp. 1929--1958, 2014.

\bibitem{rhinehart2018deep}
N.~Rhinehart, R.~McAllister, and S.~Levine, ``Deep imitative models for
  flexible inference, planning, and control,'' \emph{arXiv preprint
  arXiv:1810.06544}, 2018.

\bibitem{hecker_end--end_2018}
S.~Hecker, D.~Dai, and L.~Van~Gool, ``End-to-{End} {Learning} of {Driving}
  {Models} with {Surround}-{View} {Cameras} and {Route} {Planners},'' in
  \emph{European {Conference} on {Computer} {Vision} ({ECCV})}, 2018.

\bibitem{Codevilla_2018_ECCV}
F.~Codevilla, A.~M. Lopez, V.~Koltun, and A.~Dosovitskiy, ``On offline
  evaluation of vision-based driving models,'' in \emph{The European Conference
  on Computer Vision (ECCV)}, September 2018.

\bibitem{Liang_2018_ECCV}
X.~Liang, T.~Wang, L.~Yang, and E.~Xing, ``Cirl: Controllable imitative
  reinforcement learning for vision-based self-driving,'' in \emph{The European
  Conference on Computer Vision (ECCV)}, September 2018.

\bibitem{RobotCarDatasetIJRR}
\BIBentryALTinterwordspacing
W.~Maddern, G.~Pascoe, C.~Linegar, and P.~Newman, ``{1 Year, 1000km: The Oxford
  RobotCar Dataset},'' \emph{The International Journal of Robotics Research
  (IJRR)}, vol.~36, no.~1, pp. 3--15, 2017. [Online]. Available:
  \url{http://dx.doi.org/10.1177/0278364916679498}
\BIBentrySTDinterwordspacing

\bibitem{dagger}
S.~Ross, G.~Gordon, and D.~Bagnell, ``A reduction of imitation learning and
  structured prediction to no-regret online learning,'' in \emph{Proceedings of
  the Fourteenth International Conference on Artificial Intelligence and
  Statistics}, ser. Proceedings of Machine Learning Research, G.~Gordon,
  D.~Dunson, and M.~Dudík, Eds., vol.~15.\hskip 1em plus 0.5em minus
  0.4em\relax Fort Lauderdale, FL, USA: PMLR, 11--13 Apr 2011, pp. 627--635.

\end{thebibliography}


\end{document}